\documentclass[man]{apa7}

\usepackage[american]{babel}
\usepackage{csquotes}
\usepackage[style=apa,sortcites=true,sorting=nyt,backend=biber]{biblatex}
\usepackage{graphicx}

\DeclareLanguageMapping{american}{american-apa}
\title{Beyond the Meta: Leveraging Game Design\\ Parameters for Patch-Agnostic Esport Analitics}

\author {
    Alan Pedrassoli Chitayat,
    Florian Block,
    James Walker,
    Anders Drachen
}
\affiliation{
    University of York\, York\, UK\\
    \{alan.pchitayat; florian.block; james.walker; anders.drachen\}@york.ac.uk
}

\leftheader{}
\shorttitle{}

\begin{document}

\abstract{
Esport games comprise a sizeable fraction of the global games market, and is the fastest growing segment in games.
This has given rise to the domain of esports analytics, which uses telemetry data from games to inform players, coaches, broadcasters and other stakeholders.
Compared to traditional sports, esport titles change rapidly, in terms of mechanics as well as rules.
Due to these frequent changes to the parameters of the game, esport analytics models can have a short life-spam, a problem which is largely ignored within the literature. 
This paper extracts information from game design (i.e. patch notes) and utilises clustering techniques to propose a new form of character representation.
As a case study, a neural network model is trained to predict the number of kills in a Dota 2 match utilising this novel character representation technique.
The performance of this model is then evaluated against two distinct baselines, including conventional techniques.
Not only did the model significantly outperform the baselines in terms of accuracy (85\% AUC), but the model also maintains the accuracy in two newer iterations of the game that introduced one new character and a brand new character type.
These changes introduced to the design of the game would typically break conventional techniques that are commonly used within the literature.
Therefore, the proposed methodology for representing characters can increase the life-spam of machine learning models as well as contribute to a higher performance when compared to traditional techniques typically employed within the literature.
}

\maketitle

\section{Introduction}
Esport titles, such as \emph{League of Legends} and \emph{Dota 2}, have amassed both large audiences and player-bases (\cite{Newzoo2022}; \cite{petrovskaya2020battle}). 
Due to the competitive nature of the genre, the player community often develop so called \emph{``metas''} as explained by \textcite{kokkinakis2021metagaming}.
According to the author, \emph{metas} are naturally discovered and developed strategies for optimum ways of playing the game that are focused in determining competitive advantage available within the current parameters of the game design.
As a result, game developers are constantly updating and changing the rules of the game in order to balance the game, prevent game stagnation as well as to maintain player interest.
This is done through releasing \emph{patches} to the game, which can alter the rules and the parameters of the game \parencite{summerville2016draft}.
Those changes can be manifested by introducing new content (such as new characters), changing existing content (such as altering the duration or cost of an in-game ability), removing old content (such as discarding excising mechanics or rules) or a combination of them (such as replacing an ability for an existing character to a new ability with unique traits).

%As a result, game developers must maintain and update their games to ensure a healthy ecosystem.
%This can lead to changing core aspects of game design through frequent patches, in order to balance the many different dimensions of the game.
%Balancing is an important part of those titles, as they can develop so called ``metas'' which are naturally discovered and developed by the player community as described by \textcite{kokkinakis2021metagaming}.

%Intro has to be updated a bit to be more generally understandable. 
%Also, paper contribution comes very late - the intro right now is a lot of background, but states the problem, importance of problem, solution presented, and why that solution is useful, very late. I would start the intro with some lines about how ML and IA in esports has boomed, sector is huge/important, data is availbale, etc, but that a lot of the work in esports analytics is flawed in it being trained on specific instances of game mechanics and rules. Yesterdays model is todays trash. So, how to we mitigate this fundamental issue that is plauging an entire domain of research? In this paper, we take work on part of this prohblem, namely new characters in MOBAs ... - Anders

%some changes to address this have been made - Alan

%not enough - problem needs to be clearer and more important. Intro must bite like a snake!

As the patch changes are generally made with the intent to re-balance the game, they often incur changes to the meta, forcing competitive players to play differently and creating new game mechanics interactions.
As outlines by \textcite{demediuk2021performance}, the way in which a game is played, and the current rules set could impact machine learning models and other forms of data analysis.
This could be caused both by differences in player decision making, as well as by the new environmental state (i.e. the changes to the game's parameters themselves).
%This could have an impact in Machine Learning models and other forms of data analysis performed in the game as outlined by \textcite{demediuk2021performance}. 
As a consequence, much of the esport literature is done focusing on a limited periods of time, to reduce the impact of those changes (\cite{demediuk2019role}; \cite{katona2019time}; \cite{pedrassoli2020wards}; \cite{tot2021you}). 
However, as the game continues to change, the performance of such models may suffer, and models may need to be retrained to support a new architecture.

%Particularly when the game releases new content, such as additional characters that were not present in the game in the past.
One example that can be observed in the esport literature comes from models that include the characters present in a match.
Typically, characters are represented by their unique Character IDs (which are arbitrarily assigned numerical identifiers), which most commonly undergo a one-hot encoding or a similar variation {\cite{summerville2016draft}; \cite{makarov2018predicting}; \cite{katona2019time}; \cite{viggiato2020trouncing}; \cite{demediuk2021performance}; \cite{ringer2023time}).
In this encoding, the unique identifier of a character is represented as a vector class.
This means that if a game only has four characters, with IDs 1 through 4 respectively, character 1 can be presented as the vector \textit{(1,0,0,0)}, while character 2 would be \textit{(0,1,0,0)}, etc...
This can lead to an architectural problem where, as the number of Character IDs increases so does the number of dimensions needed to encode them.
Using the same example, if a new character is then introduced with Character ID 5, the vector needed to represent all previous character would need an additional dimension. 
Thus character 1 would then be represented as the vector \textit{(1,0,0,0\underline{,0})}.
A machine learning model developed prior to the introduction of the 5th character would be architectured to support a 4-dimensional input vector for each character, and thus not support a 5-dimensional input needed given the new game design parameters.
This could lead the model to be unusable due to the technical constraints of how it was trained and applied.

Additionally, characters within esport games usually have unique abilities and traits which allows them to be played in different ways (\cite{demediuk2019role}; \cite{demediuk2021performance}).
If a character is re-designed in a way in which several of their original abilities or traits are altered, the way in which they are played could significantly change from patch to patch.
Those changes would not be encompassed by the Character IDs.
Using the same example as above, if character 1 is designed to poses several supportive traits (such as healing or otherwise improving the in-game performance of allies) but then is re-designed to have several offensive abilities, the way in which the character is played could be changed significantly.
However, as the character ID would remain the same, a model that uses the ID could have its performance impacted, as it had been trained in a different state that does not account for the new changes to the character.
In these hypothetical scenario, while the model would still be able to produce results, it would be unclear if the results are reliable due to the uncertainty of the impact of the change to the game's environment.

Thus, this paper conjectures that changes to game design parameters may be interpreted in three ways:
\begin{itemize}
    \item Breaking changes - Those are fundamental changes to the design of the game which would require a changes in model architecture in order to produce any results.
    \item Impactful changes - Those do not incur changes in architecture, however have a substantial impact in model performance.
    \item Unimpactful changes - Those do not alter the state of the game significantly enough to affect the performance of previously trained models.
\end{itemize}
While breaking changes are usually trivial to identify, due to their severe disruption to the application of models, differentiating between impactful and unimpactful changes may require analysis.
Furthermore, if either breaking or impactful changes are identified, a previously trained model may need to undergo the training process again to account for the new parameters.
This can be a cumbersome process, especially as esport titles typically change rapidly and abruptly (\cite{summerville2016draft}; \cite{kokkinakis2021metagaming}).

This paper builds on some of the methodology suggested in the literature (\cite{demediuk2019role}; \cite{viggiato2020trouncing}), to generate a novel character representation.
This form of representing characters utilises patch specific game design data, which is then clustered to allow for a fixed and reusable notation that can be readily applied to future models.
The proposed method holds meaningful information about the character's capabilities and it is sensitive to changes introduced in patches.
%Additionally, centroid based clustering techniques (such as K-Means) allows unseen data to be represented in this format without changing the number of dimensions.
This has been done using \emph{Dota 2} - a popular Multiplayer Online Battle Arena (MOBA) esport title - which contains over 100 unique characters.
Each playable character has a range of unique abilities - which are active skills that can be used by players during play with in-game effects, such as causing damage or healing allies - and stats, including ``\textit{Intelligence}'', ``\textit{Agility}'', ``\textit{Strength}'', etc...
\emph{Dota 2} is a team based game, in which two teams of 5 players each attempt to destroy the main building in the opponent's team base.
This game has been chosen as a focus in this study due to its popularity within academia (\cite{katona2019time}; \cite{pedrassoli2020wards}; \cite{tot2021you}; \cite{semenov2017performance}; \cite{hodge2019win}; \cite{agarwala2014learning}; \cite{ringer2023time}; \cite{makarov2018predicting}), large complexity and abundant access to data.
While the resources made available within this paper are designed for \emph{Dota 2}, a similar methodology could readily be applied to other MOBA titles - such as \emph{League of Legends} - with minimal alterations, as well as advising the development of similar techniques for other esport genres.% (such as first-person shooters that could investigate the changes to properties of gun or character abilities when present).

The character representation proposed in this paper is then tested and evaluated in a case-study, which simulates a hypothetical future work within the esport literature.
Three versions of a Neural Network (NN) are trained using professional games of \emph{Dota 2} from patches 7.27 to 7.33.
These models attempts to predict the number of kills (also referred to as the score) for each team at the end of the match.
It is important to note that these NNs bear no direct contribution within the paper other than as a evaluation metric.
They serve as an illustration for possible use case within future research, and outline the capabilities of this methodology.
(NN1) was used as a simple base-line, where only the match duration was used as an input. 
No additional features were included in this baseline, therefore this network would have no way of determining or modeling the characters present in the game. 
(NN2) was utilised as an additional control, where Character IDs (one-hot encoded) were used to represent the characters present in each team, as well as the duration of the match.
This network controls for the standard encoding typically used in the literature, which holds information about the characters that have been selected in the match.
Lastly, in order to evaluate the framework proposed, (NN3) was trained using the characters selected represented through the clustered approach proposed in this paper, as well as the match duration.

Towards addressing the impact of game changes in machine learning models and other forms of data analytics within esports, which poses a problem of short life-spam, this study provides 3 major contributions: %towards solving the problem X, which is important due to Y, this paper provides Z contributions:
\begin{itemize}
    \item A feature set of character traits is compiled through the literature and game-design data made available by the game's publisher.
    \item A novel way of representing characters is proposed and validated through performance, outlying how it is sensitive to patch specific context as well as reliant to fundamental changes to the core game environment.
    \item Access to standardized format, centroids, clustered abilities and characters are made freely available for use for future research in the field\footnote{\url{https://github.com/ChitaAPC/Dota2CCR}}.
\end{itemize}

%\section{Multi-Player Online Battle Arena games}
%commenting this out for now because headers keeps confusing me when reading the PDF
%add a section where you describe MOBAs and characters in them

\section{Related work}
%when you write this, think about the reader - you have to write for someone who knows CS and maybe games, but not esports. This will dramatically increase readership. Also, consider relating the work to traditional sports to broaden the appeal of the paper)
Within the esport literature many studies which aims to apply data analysis and machine learning techniques to esport data can be found.
Those studies investigate a range of problems and applications, from classifying complex patterns \parencite{demediuk2019role}, performing micro/macro predictions about the game state (\cite{katona2019time}; \cite{hodge2019win}) or provide advisory tools to tournament organisers \parencite{tot2021you}, esport audiences \parencite{pedrassoli2020wards}, and players \parencite{summerville2016draft}.
The application of machine learning and other forms of analytics is vast within the literature.
In this section, this paper will summarize and evaluate different examples of applications, with a focus on how characters were utilised and the potential impact of game changes to the work proposed, including any steps taken to mitigate this impact.

\textcite{summerville2016draft} provides an early example of the application of machine learning techniques.
In this paper, \citeauthor{summerville2016draft} utilises both Bayes Nets (BNs) and Long Short-Term Memory Recurrent Neural Networks (LSTM RNNs) to investigate the character selection part of the game, also known as the ``\textit{Draft Phase}''.
Using those techniques, the paper attempts to predict the characters which professional players choose to play for a given matches.
In this work, the authors outline the significant impact of patches to machine learning models, due to changes to the core game mechanics that alter the environment severely and abruptly.
In this example, the characters were represented by One-Hot encoding as described previously.
This means that, as identified by \citeauthor{summerville2016draft}, their model validity is no longer guaranteed as the parameters of the game change in patches.
Additionally the model would be unable to support the addition of new characters.

\textcite{demediuk2019role} introduces clustering techniques to automatically identify player roles (which are play styles adopted by players within a team strategy) within \emph{Dota 2}.
%In this study, \citeauthor{demediuk2019role} clustered a complex range of input values to identify and classify player roles.
These roles are community determined and are not labelled within the game, but instead relate to how a player may play (for example) in a ``\textit{support}'' role, aiming to enhance their teammates rather than themselves, while other players may play in a ``\textit{carry}'' role, which aims at accumulating in-game resources to improve themselves while being supported by their teammates.
Therefore, the study focused on player performance - i.e. player decision making and how it impacts the game.
Later the authors extended this work to measure the player performance (i.e. how well a player is performing in a particular match) based on their detected roles \parencite{demediuk2021performance}.
In this extended work, the authors utilised archetype analysis - a form of unsupervised learning similar to clustering - in the Key Performance Indicators (KPIs) available in the game to provide insights into how well players are performing within their detected role.
%\citeauthor{demediuk2019role} \shortcite{demediuk2019role, demediuk2021performance} focused exclusively on player decision making data extracted from in-match events.
%While this is a complex problem which requires several insights and techniques, the author does not include contextual information about the game.
%Furthermore, the classification and measurements proposed by the authors may be impacted by changes in the meta as the game adapts to the ever evolving environment \parencite{kokkinakis2021metagaming}.
In both cases the authors utilised the order in which abilities were upgraded by the players as a key indicator of role detection (in addition to other metrics such as spatio-temporal data).
This outlines a clear importance on the abilities of the characters, and how they may dictate play-style, however, it is important to note that as patches change abilities, their effect  could also change.
This could lead to a limiting the life-spam of such model, as abilities are only represented by their ordering within the character.
To mitigate that impact, the clustering model would likely need to be re-trained for each character when \textit{impactful} changes are present as identified by the authors.

Similarly, \textcite{makarov2018predicting} proposed a model for \emph{Dota 2} (as well as \emph{Counter Strike}) that utilises roles to generate a probabilistic prediction of the winner of the match.
Within the \emph{Dota 2} model, \citeauthor{makarov2018predicting} utilised the knowledge of professional player role (as opposed to automatically detecting it through performance) in addition to other features, such as aggregate metrics of damage and resources to formulate its prediction.
This model achieved predictions ranging from \textit{73\%} to \textit{90\%} AUC depending on the input features used in the function.
Amongst the input features, the authors used the characters present in the match as one of the function's parameters.
While the exact form in which characters are used is not stated, it is conjectured that those undergo One-Hot encoded or similar.
This is conjectured as this categorical value would need to be encoded into a numerical representation in order to be utilised in the function.
For this reason, this function may also be subject to impact of changes in the game.
However, it is unclear how much weight characters have within the function, and the use of other metrics may obfuscate or otherwise help minimize the impact of such data.

Another example is the work by \textcite{hodge2019win}, in which the authors developed a Random Forest model to predicting the winner of a \emph{Dota 2} match.
In \textcite{hodge2019win}, the use of match-state is used to predict the outcome of the game.
%Similarly to the previously mentioned studies, the proposed methodology focuses on player decision making and related game state events that have happened in a particular match.
The model achieved accuracy ranges varying from \textit{70\%} to \textit{90\%} depending on the in-game time of the prediction.
This high level accuracy was achieved without including any character selection information, instead utilising only aggregate values from the match of features such as resource distribution (i.e. gold and experience), score (kills), and other in-game metrics.
The use of these aggregate features and lack of information about the characters can mitigate the impact of patches, for the potential cost of contextual information.
For example, a match which contains several characters that benefit from fighting in both teams could be interpreted differently from a match which contains several characters that benefit from avoiding fights and instead focusing on gathering/denying resources.
Furthermore, it is important to note that even though characters are not included in the model directly, the current meta could still have consequential impacts in the model.
This means that, if the meta when the model was trained revolved around team-fights and encounters, then more kills are to be expected.
A change in the meta that pivots away from confrontations (or vise-versa) could then have an impact in the performance of the model.

By contrast, \textcite{semenov2017performance} outlines a comparison of performance of different techniques for predicting the winner of a match using data from character selection alone.
Across the many methodology evaluated by the authors, it is clear that accuracy decreases as skill proficiency increases (i.e. the higher the player rank, the harder it is to predict the winner based on character selection alone).
The performance of the different methodology ranged from approximately \textit{71\%} (Factorization Machines classifier for normal skill level) to approximately \textit{64\%} (Naive Bayes for very high skill matches).
In this evaluation of models, the authors encoded the character present through a variation of One-Hot encoding, where all characters present for the Radiant team where encoded as 1, while the characters present in the Dire team were encoded as -1 and 0 otherwise.
Thus this approach is subjected to breaking changes when new characters are released (as previously explained in this paper) and would also be subjected to changes in the meta, as explained in the literature (\cite{kokkinakis2021metagaming}; \cite{summerville2016draft}).

\textcite{katona2019time} introduced a different approach to handling esport characters.
In this study \citeauthor{katona2019time} utilised a predictive deep neural network, which utilised shared weights in the inputs which pertain to a single player/character within the dataset to predict the chances of a character being killed within five seconds.
This approached allowed for greater flexibility than separating the characters, as the same weights of a network could be trained multiple times across all players for a given match.
Later this work was extended by \textcite{ringer2023time}, in which a similar approach was taken to handle the characters and their associated data.
%However, while this methodology offers some contextual information for the current character state (such as their current and max health), this is highly driven by the players' specific performance and decision making - such as itemization.
Comparably to previously mentioned works, both \citeauthor{katona2019time} and \citeauthor{ringer2023time} represent the character selection through a One-Hot encoding of the Character IDs, similarly arising the risk of breaking changes as new content is released.
Additionally, information related to the abilities of the characters were also utilised, indicating another example were key importance to the abilities can be found within the literature.

%maybe remove this paragraph completelly?
Older work in the literature by \textcite{agarwala2014learning} has attempted to utilise character selection data to identify strategies in order to predict the outcome.
In the work by \citeauthor{agarwala2014learning}, the use of aggregate performance metrics for the characters themselves (across all players) were clustered, in order to determine similarities.
This work differs from other works in the literature, as it focuses on the characters overall performance across the whole population, as opposed to individual player performance, such as (\cite{demediuk2019role}; \cite{makarov2018predicting}), or simply in the presence of the character in the match as a binary value such as \parencite{semenov2017performance}.
According to \textcite{agarwala2014learning}, using aggregate metrics of the population was insufficient to meaningfully represent the state of the game and detect strategies in order to produce accurate predictions.
However, should more refined data be used, similar approaches could be beneficial to addressing the problem of changes to the game.

Later, \textcite{viggiato2020trouncing} has taken the idea suggested by \citeauthor{agarwala2014learning} further. 
By focusing on individual player past performance with a character (i.e. wins/losses), in addition to some game design parameter about the character for a given patch (e.g. max health, attribute, attribute gain, etc..), as well as several other features, \citeauthor{viggiato2020trouncing} proposed a XGBoost model to predict the winner of the match given the character selection.
By performing feature analysis, the authors concluded that, in addition to individual player's previous experience with their chosen character, the character's ``\textit{raw attribute}'' (i.e. the game design parameters of the character excluding any player decision making) are the most significant features in predicting the outcome of a game within their model.
In this work, the authors collected over 55 thousand professional games from the years of \textit{2012} to \textit{2020}.
This data expanded many patches, thus, was subjected to many core game design changes.
The authors addressed this by parsing through the game's change logs (patch notes), which contains detailed information about the current game state at any given date.
Thus the current iteration of game for any given match could be contextualised.
However, the model was limited on character attributes without including any information about character abilities, which other points of literature suggest bear significant importance to how a character is played (\cite{demediuk2019role}; \cite{demediuk2021performance}; \cite{makarov2018predicting}; \cite{katona2019time}; \cite{ringer2023time}).
Furthermore, \citeauthor{viggiato2020trouncing} suggest that more data and a further breakdown of additional features may improve the performance further.
Additionally, the author also included the characters present in a match using through a One-Hot encoding, making it similarly susceptible to breaking changes as new characters are introduced.

Overall, several points can be extracted from the literature.
It is clear that changes in game design can impact the performance of models (\cite{summerville2016draft}; \cite{viggiato2020trouncing}).
It is also clear that high accuracy and proficiency can be achieved in models within the game iteration used during training, be it to predict the overall outcome \parencite{hodge2019win} or to perform micro prediction about in-game events \parencite{ringer2023time}.
High emphasis in the importance of character attributes and stats can also be observed \parencite{viggiato2020trouncing}, as well as the importance of abilities, which can dictate how characters are played (\cite{demediuk2019role}; \cite{demediuk2021performance}; \cite{makarov2018predicting}; \cite{katona2019time}; \cite{ringer2023time}).
It is also clear that character selection is a very common feature utilised in the literature (\cite{summerville2016draft}; \cite{makarov2018predicting}; \cite{katona2019time}; \cite{demediuk2019role}; \cite{viggiato2020trouncing}; \cite{demediuk2021performance}; \cite{ringer2023time}).
However, the typical approach to including character selection subjects models to breaking changes when new characters are introduced, or to impactful changes through updates to existing game design parameters.
This could lead to the need of frequent retraining models, which can be a cumbersome process.
Despite its wide spread utilisation, very little work has been done to formulate a consistent form of character representation that is patch-aware, to allow for changes in the game design parameters to be included in the analysis while supporting the addition of new characters.
This paper aims to address this gap, by building from \parencite{demediuk2019role} and \parencite{viggiato2020trouncing} to introduce a methodology that allows researchers to contextualise game design parameters in a robust way that can support new characters without compromising the flow of data.
Through analysing patch notes, it is possible to support a more rich environment for representing characters that includes data from both, their attributes and their abilities.

\section{Methodology}

This section describes the steps taken to collect and process the data used.
As explained above, the methodology proposed focuses on the \emph{Dota 2} title, as a popular esport game with a wealth of academic research, and large player-base/audience.

\subsection{Data Collection}
In order to conduct this study, two distinct datasets were collected.
Firstly patch data was collected, which consists of data about game design parameters for characters, including of the abilities and attributes from patches 7.27 to 7.33.
%This was used to compile Dataset (1).
Secondly, a match history dataset was compiled which consists of the character selection data of all competitive matches for these patches, as well as the duration of the match, the number of kills on each side and the result of the match.
%This constituted Dataset (2).

Both datasets were collected through the OpenDota API\footnote{\url{https://docs.opendota.com/}} and its associated public table of constants\footnote{\url{https://github.com/odota/dotaconstants/tree/master/build}}.
OpenDota is a free platform that offers in-depth statistics and break down of \emph{Dota 2} public matches, including those in professional tournaments and events.
This platform has been commonly used by other works in the literature for data collection (\cite{hodge2019win}; \cite{demediuk2019role}; \cite{demediuk2021performance}).

%- i.e. a set of game values which are considered constants by the game's developers for any given patch -  
As OpenDota's table of constants is stored in a Git repository, it is possible to access older versions of the tables, referring to previous game patches, through the history of individual files or folders.
The first dataset was compiled by parsing through three JSON files retrieved from this repository (``hero\_abilities.json'', ``abilities.json'' and ``heroes.json'').

The second dataset was compiled through OpenDota's SQL query feature available on their platform.
The data collected included:
\begin{itemize} 
    \item MatchID - a unique identifier for the match
    \item Patch - the patch in which the match was played in (such as '7.27')
    \item Duration - The duration of the match in seconds
    \item KillsR - The number of kills obtained by the Radiant team
    \item KillsD - The number of kills obtained by the Dire team
    \item HeroX - The hero played by player X, where X is a number from 0 to 9. Players 0-4 correspond to the Radiant team, and 5-9 correspond to the Dire Team
    \item RadiantWin - A binary variable containing 1 if Radiant won the game and 0 if Radiant lost (note it is not possible for a match to end in a draw in \emph{Dota 2})
\end{itemize}

Only professional and premium matches were collected from patches \textit{7.27} to \textit{7.33}, which included data deom \textit{Jun 2020} to present.
Games which did not conclude in a natural state (for example if a player has abandoned, or if there had been a server error) were not collected.
This lead to a total of \textit{61,254} matches from the 7 different patches. (note that at the time of writing, the latest patch (7.33) is still active, and therefore fewer matches are available. 
% + 2235 from patch 7.33; original 59,019

\subsection{Data Processing}
In order to perform clustering on the game design data, the JSON files retrieved from the table of constants were processed.
Firstly ``hero\_abilities.json'' contains the name of each individual character and their associated abilities.
This file was parsed to compile a list of the relevant character abilities, which allowed abilities from non-playable entities, such as neutral characters, to be discarded.
Using the name of each individual ability as a key, ``abilities.json'' was then parsed to generate a CSV file containing the properties of every character ability for each of the relevant patches.
This was achieved by compiling a script that normalise and standardise the way data is represented.

Several inconsistencies were detected in the way data was stored by the game between patches and abilities.
For example, an ability with the property of ``movement\_speed\_slow'' of \textit{30} and another ability of ``movement\_speed\_bonus'' of \mbox{\textit{-30}} both have the same effect of reduce the movement speed of the target by \textit{30} units.
Through manually comparing the properties names and values and cross referencing to the game's wiki, a script that standardised the properties was compiled.
While this approach is subject to errors, gathering data is limited by the way in which data is made available by the game's publishers.
%It is important to note that this does not fully encapsulate all the nuances of \emph{Dota 2}, due to the large complexity of the game.
However, extracting data using this methodology allows for more in-depth analysis than previous techniques which generally ignore what abilities do and their associated properties (\cite{demediuk2019role}; \cite{katona2019time}).
%Therefore, while some impact in the analysis is expected, it would be unfeasible to fully represent the entirety of the problem space at this stage.
%Additionally, this data will be used to perform clustering, which by its nature will further obfuscate data.
Once all of the properties for every ability was compiled, info on the characters attributes - as described in the literature \parencite{viggiato2020trouncing} - was extracted from the ``heroes.json'' file and amended to the relevant abilities for the heroes.
Thus, each entry in the newly compiled CSV file contained data for each ability for each character as well as data about the character itself, which contextualises its use in the game.
The CSV files have been made available in the resources listed in the Introduction section for future use.

\subsection{Cluster Analysis}
Once all of the character abilities were processed and the CSV files were compiled it was possible to utilise clustering algorithms on the data.
As this study aims to provide longevity to machine learning models in esports analytics and support the addition of new unseen data, K-Means \parencite{humaira2020determining} was utilised as it can be readily used without altering the labeling or number of clusters once the centroids are found.
This means that once the centroids are found, the model would not need to be retrained with unseen patches.
Instead, the data from new patches could simply be predicted with the existing centroids, therefore clusters do not change when adding new data.
This approach would then contribute to reducing the impact of breaking changes, as the value of K would not need to change unless the new data leads to a new concentration of clusters (i.e. an impactful change).
In this case analysis of the cluster distribution - such as that described in this section - would reveal the need for new clustered.
However, data from several patches are used to generate the clusters to reduce the risk of impactful changes being present.

In this study, K-Means was used and the ``Sum of the Squared Error'' (SSE) was used to evaluate the split of the data and identify an optimum value for K \parencite{humaira2020determining}.
Figure~\ref{fig:kmeans_sse} outlines the elbow plot generated with different values for K.
As the figure outlines, the elbow lies roughly between \textit{K=40} and \textit{K=75}.
The exact value for K was then selected using the highest ``Silhouette Score'' \parencite{humaira2020determining} within this range, which provided \textit{\textbf{K=68}}.

\begin{figure}[t]
\centering
\caption{The Elbow Plot for the K-Means clustering of the character's abilities and traits together}
\label{fig:kmeans_sse}
\includegraphics[width=1\columnwidth]{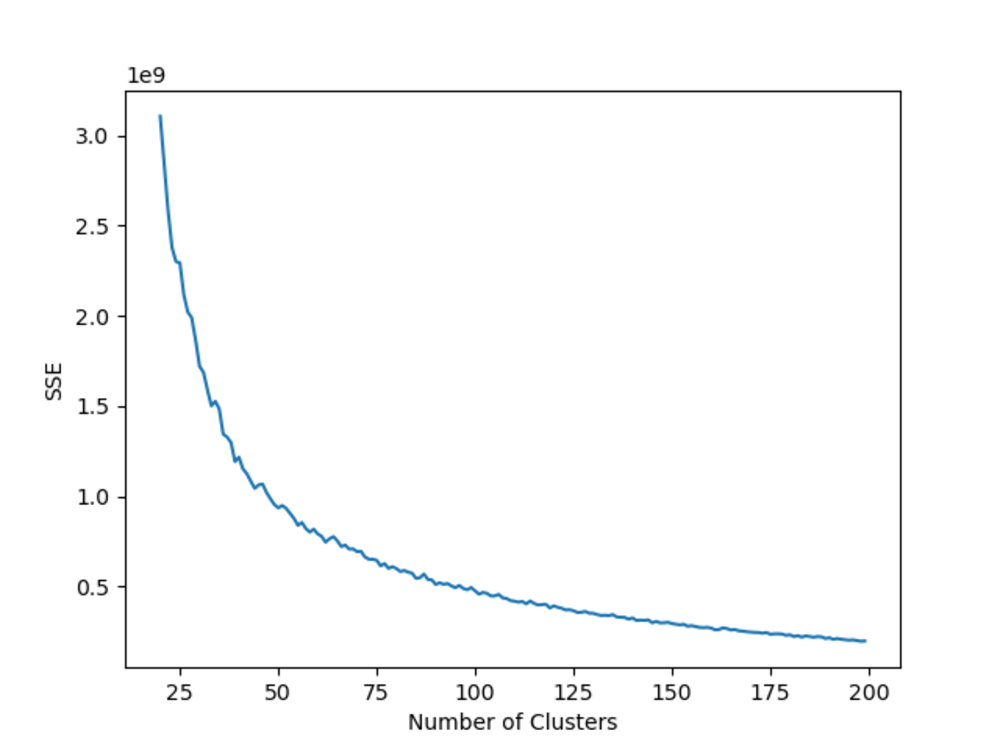}
\end{figure}

Once clustering was performed with the ability data, the labels obtained could then be used to represent characters.
As one character contains multiple abilities, the cluster for each ability was encoding into a one-hot-encoded format.
This means that any single ability was then represented as a K-long vectors of zeroes, except for the \textit{i}'s dimension, where \textit{i} is the cluster number for that particular ability.
Subsequently, one character can then be represented by adding all of the K-long vectors of their abilities.
This allows one character to be represented as a K-long vector, containing the modes for each of their ability clusters.
Unlike the One-Hot encoding of Character IDs, however, this approach would not be subjected to breaking changes when new characters are introduced, as the value for K would remain the same.

For demonstrations purpose, this approach can be applied to a fictional game, where K=3.
In this fictional game ``Character X'' has three abilities, of clusters 0, 0 and 2 respectively.
This means that Character X can be represented as the addition of the vectors (1, 0, 0); (1, 0, 0) and (0, 0, 1).
Through this technique, Character X can then be represented as a single K-dimensional vector of (2, 0, 1) - where the mode of \textit{cluster 0} is two, \textit{cluster 1} is zero and \textit{cluster 2} is one.
Conversely, a more complex character with 5 abilities of clusters 0, 0, 1, 1 and 2 respectively could be represented as the single vector (2, 2, 1).
This simplistic, yet powerful approach allows for complex patterns to emerge from the data while still maintaining the meaning of each cluster as identified by the data.
Furthermore, it maintains the size of the vector needed to represent any characters to K, even when new abilities or new characters are introduced by future patches.

\subsection{Predictive Neural Network}
Three NN are used to help evaluate the character representation.
This simulates a potential use-case within future research, with two baselines.
The case-study of predicting the number of kills at the end of a match was selected, as this is a under-explored area in the literature with some of the potential highlighted by existing research (\cite{viggiato2020trouncing}; \cite{tot2021you}).
This is also a similar use-case as to predicting the winner of a game, which contains a wealth of knowledge (\cite{viggiato2020trouncing}; \cite{agarwala2014learning}; \cite{hodge2019win}; \cite{makarov2018predicting}; \cite{summerville2016draft}). 

Firstly (NN1) was trained, which used only the match duration and no additional information.
This provided a baseline for comparison, as all models include the match duration.
This was included as the literature suggests that duration bears a significant impact in the outcome of the game, both in terms of result and in terms of score \parencite{viggiato2020trouncing}.
Secondly (NN2) was trained which contained both match duration and character selection data.
The character selection data was represented by encoding the Character IDs into a one-hot encoding as a standard technique in the literature (\cite{semenov2017performance}; \cite{ringer2023time}; \cite{pedrassoli2020wards}).
Lastly, (NN3) was trained using only the duration and the character selection data as represented by the clustered approach.
In order to train this model, whole line-ups - rather than individual characters - were used.
This was achieved by replicating the steps for transforming an ability vector into a character vector as depicted in the previous section.
The individual character vectors for a team's selection is added together collapsing it into one K-long vector.
This means that vectors for abilities, characters and team compositions can all be represented in the same number of dimensions, which expresses the modes for each cluster available.

All matches for patches \textit{7.32 \& 7.33} were excluded from the training set to be used as special test comparisons.
Patch \textit{7.32} introduced a brand new character, \emph{Muerta}\footnote{\url{https://www.dota2.com/hero/muerta}}.
Patch \textit{7.33} had several major changes to the game, including the way in which characters primary attributes are handled.
In this patch, a new type of primary attribute was created that makes no attribute the primary. 
This means that - where before, characters were classed as one of ``\textit{Strength}'', ``\textit{Agility}'' and ``\textit{Intelligence}'', as of the latest patch characters can also be ``\textit{Universal}''\footnote{\url{https://dota2.fandom.com/wiki/Universal}}, which means no attribute is their primary attribute.
The data collected for the two patches serve as a illustration of breaking and impacful changes.
Additionally, the remaining match data (patch \textit{7.27} to \textit{7.31}) were split using a (64/16/20)\% split into a training, test and validation datasets.
Table~\ref{tab:nn_dataset_sizes} outlines the number of matches per each of the different data splits.

\begin{table}[h]
\centering
\caption{Number of matches per split used for training and evaluating the neural networks from the total \textit{61,254} matches in the dataset}
\label{tab:nn_dataset_sizes}
\begin{tabular}{|c|c|c|c|c|}
    \hline
    \textbf{Train} & \textbf{Validation} & \textbf{Test} & \textbf{Test} (7.32) & \textbf{Test} (7.33)\\
    \hline
    \textit{27,171} & \textit{6,793} & \textit{8,491} & \textit{16,564} & \textit{2,235}\\
    \hline
\end{tabular}

\end{table}

All networks were trained using the same architecture with the exception of the input layer, as described above.
The networks consist of 6 hidden layers with (\textit{1024, 512, 128, 64, 32, 8}) neurons respectively, with the ``Sigmoid'' activation function.
These model predicts the number of kills for each team simultaneously (i.e. two outputs neurons, one for each team).
Because of the simplistic architecture and the amount of data, batching was not used (i.e. batch size of 1) and the model was trained for 100 epochs, which was sufficient to outline the trends, detailed in the Results and Discussion Sessions.
The networks were trained with an \textit{Adam} optimizer and learning rate was set to \textit{1e-4}.
No dropout layers or regularizer were used.
Figure~\ref{fig:nn_train} Outlines the train and validations loss for each of the three networks per epoch.
Further work to this architecture could produce more reliable results, however a comprehensive study on this use-case is beyond the scope of this paper.

\begin{figure*}[t]
    \centering
    \caption{Training and validation loss and AUC graph per epoch for each neural network.}
    \label{fig:nn_train}
    \includegraphics[width=1\textwidth]{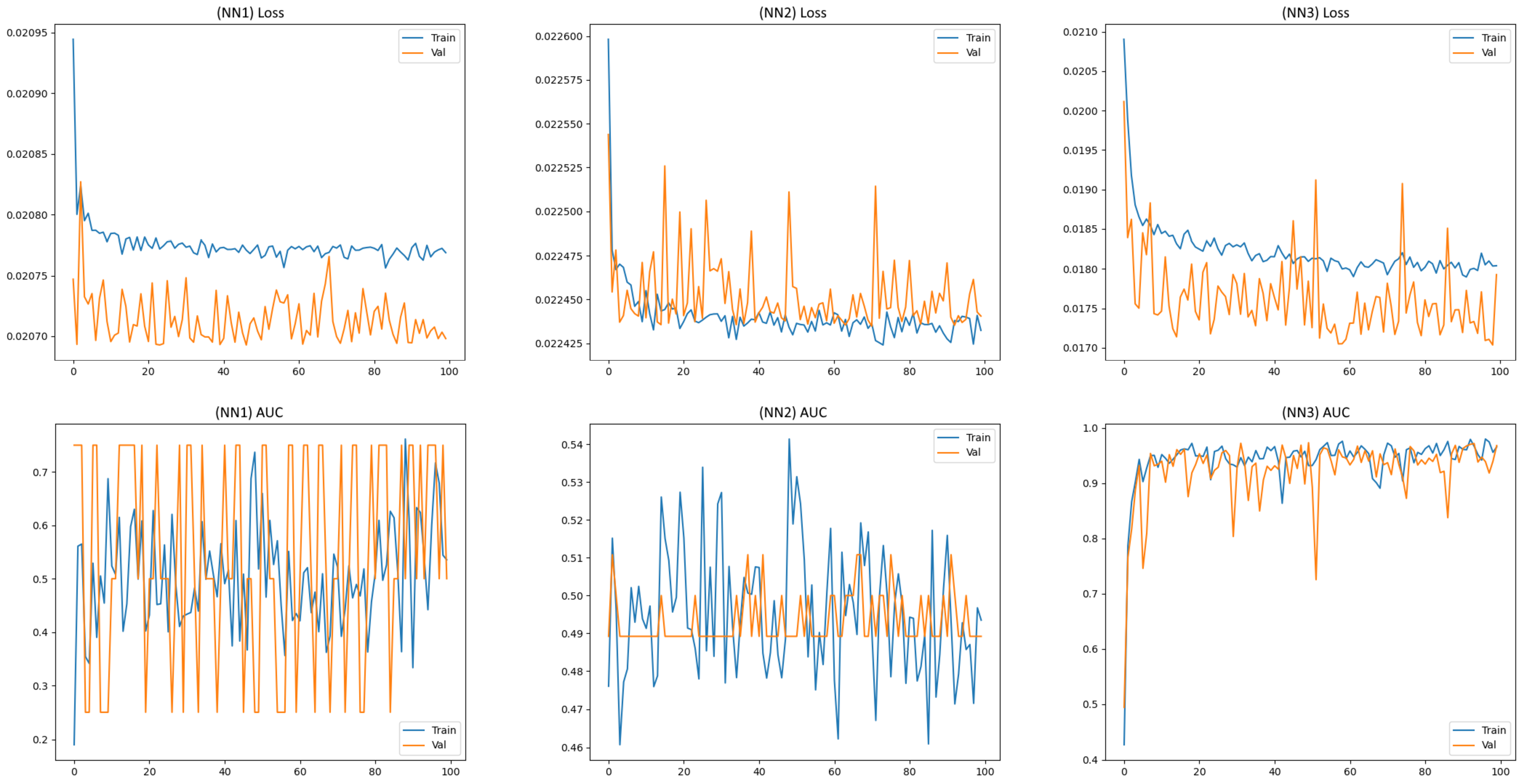}
\end{figure*}

The performance of the predictions were then compared.
This includes the training, validation and test performances for patches from 7.27 to 7.31, which outlines the performance within representative data.
Additionally, the performance of predictions for games in patch 7.32 and 7.33, which introduced the new character and new character types were also compared.
In order to make those predictions possible for (NN2), the maximum value used for the Character ID encoded included the Character ID for \textit{Muerta}.
This was made with the sole purpose of ensuring the architecture supports the additional dimension.
It is important to note, however, that this dimension would - necessarily - always have zero in the input vector for all training data.
Therefore this approach would not be suitable for actually training models with real world applications and it is limited to exactly one additional character (thus, any future characters released would not be supported by the (NN2)'s architecture).

\section{Results}
The results obtained for all networks can be found in Table~\ref{tab:nn_auc}.
The test dataset contains unseen matches from patches 7.27 to 7.31 - which are the same patches used in training dataset.
Conversely, all matches from Patch 7.32 and 7.33 have been entirely excluded from the training dataset.
Patch 7.32 introduced a new character, while Patch 7.33 (the currently active patch as of time of writing) introduced some significant changes to the game design, such as a new character type, as previously discussed.
%This means that all three networks were entirely trained without any data from the new character and character type.

\begin{table}[h]
\centering
\caption{AUC of the kills prediction for each dataset}
\label{tab:nn_auc}
\begin{tabular}{|l|c|c|c|}
    \hline
    \textbf{NN} & \textbf{Test} & \textbf{Test} (7.32) & \textbf{Test} (7.33) \\
    \hline
    (NN1) & 0.50 & 0.50 & 0.50\\
    (NN2) & 0.50 & 0.49 & 0.49\\
    (NN3) & 0.85 & 0.85 & 0.86\\
    \hline
\end{tabular}
\end{table}

Furthermore, both (NN1) and (NN2) typically produced the same prediction value for both outputs.
This means that for most matches, the prediction for the number of Radiant kills was exactly equal to the prediction for the number of Dire kills.
Between the approximately 27 thousand matches in all test datasets (Test, Patch 7.32 \& 7.33), (NN1) only produced different predictions for 351 matches while (NN2) produced different prediction 403 of the matches.
On those cases (for both networks), the difference in kills only varied by one between teams.
This was not the case for (NN3), which only predicted the same number of kills in approximately 3 thousand matches (11.1\% of the dataset).

\section{Discussion}
As outlined by Figure~\ref{fig:nn_train}, all three models had a sharp decrease on the loss followed by a plateau within the first 25 epochs.
This suggests that all three models reached their convergent point within a few epochs.
It is also noteworthy that the train and validation losses did not deviate significantly at any stage for any of the three neural networks.
This suggests that none of the models overfit to the training data.
Table~\ref{tab:nn_auc} also suggests that there was no overfitting in any of the models, as the overall performance of each network was consistent from that observed with its respective training/validation accuracy at the 100th epoch.

Furthermore, both (NN1) and (NN2) produced very similar results throughout all test datasets.
Thus, it is conjectured that (NN2) trained in a way to put very little significance for the lineup vectors, instead having more significant weights around the duration feature.
This brings to question the use of line-up vectors as input features.
Particularly when comparing similar performance in literature models observed when comparing the works of \textcite{hodge2019win} - which does not include any character selection data - and \textcite{summerville2016draft} - which does - as well as the comparatively lower accuracy obtained by \parencite{semenov2017performance} which only uses character IDs to \parencite{viggiato2020trouncing} which includes additional information.

Moreover, both (NN1) and (NN2) produced a relatively flat training AUC graph, despite the drop in loss at the early epochs.
This may indicate that the models trained to produce a consistent prediction that minimizes the loss without detecting any meaningful patterns.
In the other hand, (NN3)'s AUC graph produces a sharp rise in AUC, which is consistent with the drop in the loss.
In addition to this, it significantly outperformed both baseline models (NN1 \& NN2), pointing to the network detecting meaningful patterns in input data.

When comparing the performance of the three networks across the Test, Patches 7.32 and 7.33 datasets, it is clear that (NN3) produced consistent results despite the introduction of the new character and the change in character primary attributes observed in the two new patches.
No observable difference could be detected in (NN1), which is expected, as it was not trained with any character information.
Although there was a small drop in performance for (NN2), this could fall within the margin of error, particularly as the predicted values for this model resembles that of (NN1).
Similarly, while there was a small increase in accuracy for (NN3) in Patch 7.33, that could also fall within the margin of error, particularly because there are fewer matches in that patch as the current live patch at the time of writing.
However, as (NN3) maintained a high degree of accuracy through all patches, that is indicative that the clustered character representation approach may hold significant underlying principles of the game that is able to contribute the prediction.

It is important to note, however, that significant changes to the core game design - beyond that of the patches present in this study - could impact the cluster themselves and change the density distribution of the clusters.
In this case new clustering may be required, which would similarly cause breaking changes to (NN3) - as the number of clusters or the meaning of a individual cluster may change.
Thus, while this approach is a step towards more robust machine learning models, it is not completely devoid of risks of future impacts.
Nevertheless, it reduces the risk of breaking changes, which provide improvements over the existing methodology (One-Hot encoding of Character IDs).

Lastly, the advantages observed in this study (maintaining high performance, supporting the addition of any number of characters) suggest that the clustered approach for representing characters can increase the longevity of models developed in future research.
Additionally, the results presented here suggests that this approach can more easily represent underlying patterns in the data, as observed by the differences in results between (NN2) and (NN3).

% Thus is it clear that the game design parameters of a given patch can be represented within the clustering approach.
% This means that \textbf{RQ1} can be addressed with the characters' ability traits as well as their stats, (as described from the literature) which is then derived from the game's patch notes.
% \textbf{RQ2} is then addressed by clustering each of the character's traits and encoding them in a K-long vector.
% Using simple vector addition, those traits can then be collapsed together to produce a complete character representation that is consistent in dimensions and through past and future patches.
% This then helps to build flexibility into machine learning models in the future of esports analytics.

%w have rto buuild flexuibility into our ML models in the future in esports 

\section{Conclusion}

In conclusion, this paper outlined how the use of character selection data is a common feature for a range of machine learning models within the esport analytics literature.
The impact risk of changes in the game design has also been covered in the literature, which often reduces the lifespam of models, limiting them to a single or few iterations of the game, while iterations are released rapidly.
Despite that, very little work has been done to support changes to the game design parameters.
In the current literature that relies on character selection data, no model could be found that decisively support the addition of a new character without needing to change its architecture (i.e. the introduction of new characters would incur breaking changes).
Through clustering characters abilities and attributes, as extracted from the literature, this paper proposes a novel character representation that both encapsulate the iteration of the game for any given patch and support the inclusion of future characters without the need for a change in architecture.

By evaluating the character representation through a case-study, this paper simulated a potential use case for future research.
Through comparing three predictive neural networks (NN1-3), this paper conclude that this novel character representation can add longevity to future models, while providing more information about the character selection than the standard methodology currently employed within the literature.
This paper has made the the resources available to use for future research, including the unclustered data; the centroids used; and the clustered representations of abilities and characters for \emph{Dota 2}.

%we need a small section on ethics of data collection - need to talk about this.
\subsection*{Ethical Statement}
In esports analytics research, a considerable amount of publicly available data is utilized. 
Such data is usually shared via APIs developed from the publisher of the game in question. 
%The data are highly important to the esports community, forming the basis for analysis of your own performance as a player or a team, is used by streamers, websites and even professional services in the space. 
%
OpenDOTA collects a significant amount of data from players of \emph{Dota 2} via querying the API system provided by the publisher of the game, Valve. 
All matches of \emph{Dota 2} played are recorded and available via the game's API, which can includes recording of gamer tags. 
No device information or similar is made available through the API, and neither are private chat logs. 
When you install \emph{Dota 2}, you agree to the data collection and sharing through the End User Licence Agreement (EULA) presented when installing Steam, the platform through which \emph{Dota 2} is installed.
While performance data is always recorded as agreed by the EULA, gamer tag is not collected unless explicitly opted in by players to agree to have this data made publicly available.

The data used in the current study does not include personally identifiable information (PII). 
Gamer tag information was not included from the dataset, as these can sometimes be used across multiple online profile accounts and thus potentially be used to identify people.  
Collectively, we consider this use of data within reasonable ethical use under research ethics norms and expectations. 

\subsection*{Acknowledgements}
This work was supported by the EPSRC Centre for Doctoral Training in Intelligent Games \& Games Intelligence (IGGI) EP/S022325/1.

\printbibliography

\end{document}